\documentclass[conference]{IEEEtran}
\usepackage{graphicx}
\usepackage{multicol, blindtext}
\usepackage{times}
\usepackage{amsmath}
\usepackage{subcaption}
\DeclareMathOperator*{\argmax}{argmax}
\DeclareMathOperator*{\argmin}{argmin}
\newcommand{\norm}[1]{\left\lVert#1\right\rVert}

\usepackage[numbers]{natbib}
\usepackage{multicol}
\usepackage[bookmarks=true]{hyperref}
\usepackage{float}

\begin{document}

\title{Smoothing and Mapping using Multiple Robots}

\author{Karthik Paga (kpaga), Joe Phaneuf (jphaneuf), Adam Driscoll (jdriscol), David Evans (dje1) @cs.cmu.edu}

\maketitle
\thispagestyle{plain}
\pagestyle{plain}

\begin{abstract}
Mapping expansive regions is an arduous and often times incomplete when performed by a single agent. In this paper we illustrate an extension of \texttt{Full SLAM} \cite{Dellaert06ijrr} and \cite{dong},  which ensures smooth maps with loop-closure for multi-robot settings. The current development and the associated mathematical formulation ensure without loss of generality the applicability of full bundle adjustment approach for multiple robots operating in relatively static environments. We illustrate the efficacy of this system by presenting relevant results from experiments performed in an indoor setting. In addition to end-to-end description of the pipeline for performing smoothing and mapping \texttt{SAM} with a fleet of robots, we discuss a one-time prior estimation technique that ensures the incremental concatenation of measurements from respective robots in order to generate one smooth global map - thus emulating large scale mapping with single robot. Along with an interpretation of the non-linear estimates, we present necessary implementation details for adopting this SAM system.

\end{abstract}

\IEEEpeerreviewmaketitle
 
\section{Introduction}

Maps are becoming increasingly important for robotics systems.  Maps are used for localizing a robot's position, planning the robot's behavior, or just communicating results to an end user.  Maps however can often be limited to only the data collected by a single robot over a small area.  By using multiple robots or a single robot across many time periods, a larger map can be generated.  These larger maps can allow for the robot to operate over a larger area or use the additional information to make more intelligent plans.

This ability to generate large maps is particularly relevant to the self-driving car field.  For a self-driving car to operate throughout a city, it must have an accurate map of the city.  It is infeasible to have a single vehicle navigate and map the entire city. Instead, by using a fleet of vehicles, a map of the city can be generated more efficiently, saving the self-driving company time and logistical complexity of having a single vehicle navigate the city.

Another relevant application of multi-robot mapping is in the field of augmented reality.  By having multiple people able to map an environment, augmented reality companies can generate larger maps that can be used to locate other users relative to each other and use this information to create a more immersive environment.

The goal for this project is to apply multi-robot mapping concepts by using multiple robots to generate a single global map. We implemented multi-robot smoothing and mapping (SAM) by performing an extension of full SLAM.  We use concepts from multi-robot SAM papers to estimate priors of robots in a global frame and solve a global full SLAM graph.  We implement our algorithm on an autonomous ground robot to map out the first floor of Newell-Simon Hall (NSH).

\section{Literature Review}
While there have been multiple contributions to the least squares optimization and particularly determining the complete trajectory of a mobile robot online and offline, for the purposes of this project we rely on the seminal work of Dellaert and Kaess \cite{Dellaert06ijrr} to perform full SLAM.

This paper highlights a tractable method to solve the full SLAM problem i.e., not just the current robot location but the entire robot trajectory and landmark positions up to the current time. 
This is achieved by structuring the problem as a least-squares optimization with emphasis on sparse matrix factorization. 

$$\delta^* = \underset{\delta}{argmin}\left \| A\delta - b \right \|_2^2$$
Over a period of time, the matrices A and b can grow to be very large but quite sparse.

To handle the large matrices, the authors discuss implementing generic approaches such as Cholesky or LDL decomposition and QR factorization are perform the optimization more efficiently.

To handle non-linear functions, the paper also highlights the ability to linearize about a known robot pose (prior) and perform non-linear optimization.

Dong et al (2015) \cite{dong} discuss an approach for multi-robot pose SLAM. They use methods based on expectation maximization to optimize for poses of several robots.  For their feature detection, the team uses 2D laser scans to compare relative poses of the robots.  

A important insight made by \cite{dong} was the need for an accurate initial pose estimates for each robot.  Without a good initial guess, then the optimizer would produce poor final results.  To overcome this challenge, the team created several initial guesses for each robot pose and allowed the optimizer to optimize for the best hypothesis.

Andersson and Nygards (2008) \cite{csam} present an approach to perform multi-robot smoothing and mapping using $\sqrt{SAM}$ techniques.  Their approach generates a single global map from several local maps.  They generate the global map by comparing measurements of shared landmarks between robots.

These corresponding measurements, also called "rendezvous" points are used to add additional constraints from a robot's trajectory to a base node.  The base node represents a global frame in which the final trajectories and landmarks will be represented in.  These constraints are added to the A and b matrices of the least-squares representation, and a final map of the landmarks and robot trajectories is calculated.

Another approach to multi-robot smoothing and mapping is detailed by Cunningham et al \cite{ddf-sam}.  Instead of using all robot measurements and trajectories to generate a single global map, their approach distributes the smoothing and mapping solver across every robot.

Each robot generates a local map from its own trajectory and receives a neighborhood map consisting of the local maps of all nearby robots.  From here, the approach sets transform constraints on the the robot's local map to the neighborhood map.  With these constraints, full SLAM is performed again to generate a single global map in each individual robot's frame.

Our approach draws heavily on techniques drawn from \cite{ddf-sam} and loosely draw from \cite{csam}.
\section{Approach}
\subsection{Overview}
\begin{figure*}[ht]
    \centering
    \includegraphics[width=\linewidth]{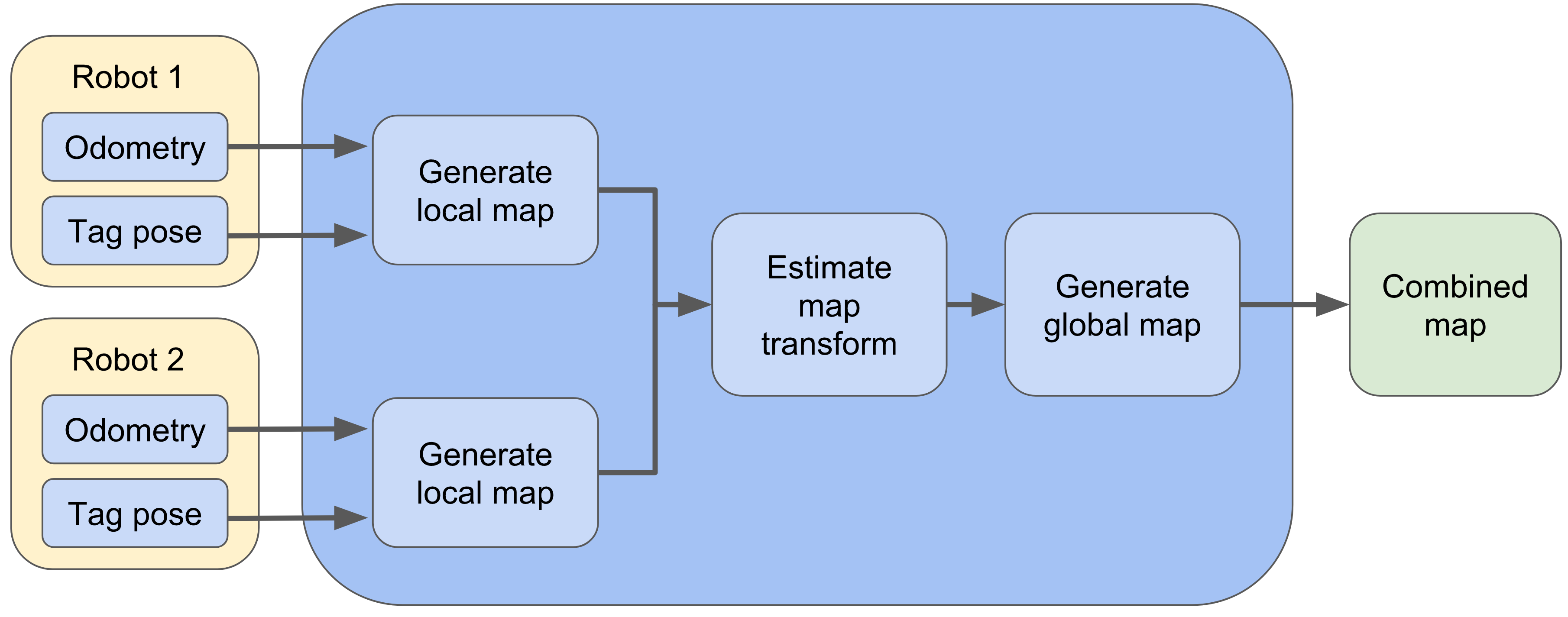}
    \caption{Functional architecture of the system}
    \label{fig: func_arc}
\end{figure*}

Before we look at how to build a map for multiple robots, let's look at at single robot mapping techniques from \cite{Dellaert06ijrr}. The goal for single robot full SLAM is to use control and sensor measurements to generate a map of the robot's trajectory and the enviornment. Dellaert and Kaees show this problem can be represented by a maximium a posteriori estimate (equation \ref{eq:maxpost}).
\begin{equation}
\label{eq:maxpost}
\Theta^* = \argmax_\Theta P(X,L|Z)
\end{equation}
Where $\Theta=(X,L)$ consists of: $X$, the trajectory of states of the robot, and $L$, the landmark positions that make up the map. Lastly, $Z$ represents the sensor measurements and control inputs.

Now, to generalize this for multiple robots in the environment we can redefine our trajectory $X$ as $\mathcal{X^R}$ which represents all the trajectories of all robots in the environment, $\mathcal{R} = \{1,...,R\}$, yielding equation \ref{eq:maxpostrob}.
\begin{equation}
\label{eq:maxpostrob}
\Theta^* = \argmax_\Theta P(\mathcal{X^R},L|Z)
\end{equation}

This can be turned into a least-squares problem (equation \ref{eq:generic_lsq}).
\begin{equation}
\label{eq:generic_lsq}
\begin{matrix}
\Theta^* = \argmin_\Theta
\{\sum_{r=1}^{R} 
\{\sum_{i=1}^{M} \norm{f_i(x_{i-1}^r, u_i^r)-x_i^r}^2+ \\ \\
\sum_{k=1}^{K} \norm{h_k(x_{ik}^r, l_{jk})^r-z_k^r}^2\}\}
\end{matrix}
\end{equation}

This is a nonlinear function, so it can be linearized by taking linear approximations of the motion models $f_i^r$ and the measurements models $h_k^r$.  We linearize the models using Taylor series approximations and then optimize function to find the min using non-linear optimization functions such as Gauss-Newton or Levenberg-Marquardt algorithms.  Because of the increased state space, we will use $\sqrt{SAM}$ techniques demonstrated by \cite{Dellaert06ijrr} to improve the computational efficiency.

This approach, however, assumes the availability of a prior for each robot.  This is a strong assumption because one does not often know the precise starting positions of the robots in the same frame.  Ultimately, we redact the assumption on the prior of each robot by estimating relative transformations between each robot's local map. These transforms can be estimated using RANSAC on corresponding landmarks, correspondences, between respective local maps. Once the priors are estimated, a final bundle adjustment is performed to merge the local maps into a final global map.

In the following sections we explain in detail the steps necessary to generate a local map, estimate the transform for the local map, and perform the final bundle adjustment to estimate one smooth global map.

\subsection{State Vector}
The output of this system will be a state vector ( equation \ref {eq:statevector} ) containing a pose at each timestep $i$ ( equation \ref{eq:odomvector} ) and a landmark position ( equation \ref{eq:lmvector} ) for each of m landmarks.

\begin{equation} \label{eq:odomvector}
x_{i}=
\begin{bmatrix}
r_x\\
r_y\\
r_\theta\\
\end{bmatrix}
\end{equation}

\begin{equation} \label{eq:lmvector}
l_{m}=
\begin{bmatrix}
l_{x}\\
l_{y}\\
\end{bmatrix}
\end{equation}

\begin{equation} \label{eq:statevector}
X =
\begin{bmatrix}
x_{i} \forall i \\
l_{m} \forall m
\end{bmatrix}
\end{equation}

\subsection{Motion model}

The motion model is designed for a differential drive robot, Groundsbot - refer section \ref{implementation}. Odometry is captured as left and right wheel linear velocities. The SLAM solver will incorporate odometry information by computing errors between predicted and measured changes in robot pose. Consequently, each odometry reading will be converted to a change in robot state. For right and left wheel velocities $V_{R}$ and $V_{L}$ and wheel base length $l$, the corresponding changes in robot state from one pose to the next are described in equation \ref{eq:motion1}. Note that $r_\theta$ in equation \ref{eq:motion1} comes from the current state estimate within the iterative solver.

\begin{equation} \label{eq:motion1}
\begin{bmatrix}
\Delta r_X \\ \\
\Delta r_Y \\ \\
\Delta r_\theta
\end{bmatrix}
=
\begin{bmatrix}
\frac { V_{R} + V_{L} } {2} cos ( r_\theta ) \\ \\
\frac { V_{R} + V_{L} } {2} sin ( r_\theta ) \\ \\
\frac { V_{R} + V_{L} } {l} \\
\end{bmatrix}
\end{equation}

Considering the relevant subset of the state vector shown in equation \ref{eq:statevectorodom} , equation \ref{eq:odompred} shows the odometry measurement prediction function at timestep i. Equation \ref{eq:odomjac} shows the Jacobian of the odometry measurement prediction function with respect to the state vector.
\begin{equation}
x=
\label{eq:statevectorodom}
\begin{bmatrix}
r_{xi-1} \\
r_{yi-1} \\
r_{\theta i-1}  \\
r_{xi}   \\
r_{yi} \\
r_{\theta i}
\end{bmatrix}
\end{equation}

\begin{equation}
h_{i} =
\begin{bmatrix}
\label{eq:odompred}
\Delta x \\
\Delta y \\
\Delta \theta
\end{bmatrix}
=
\begin{bmatrix}
r_{xi} - r_{xi-1} \\
r_{yi} - r_{yi-1} \\
r_{\theta i} - r_{\theta i - 1}
\end{bmatrix}
\end{equation}

\begin{equation}
\label{eq:odomjac}
H_{i} =
\frac { \partial h_{i} } { \partial x } =
\begin{bmatrix}
-1 &  0 &  0 & 1 & 0 & 0 \\
 0 & -1 &  0 & 0 & 1 & 0 \\
 0 &  0 & -1 & 0 & 0 & 1  \\
\end{bmatrix}
\end{equation}

\subsection{Sensor model}
\begin{figure}
    \centering
    \includegraphics[width=\linewidth]{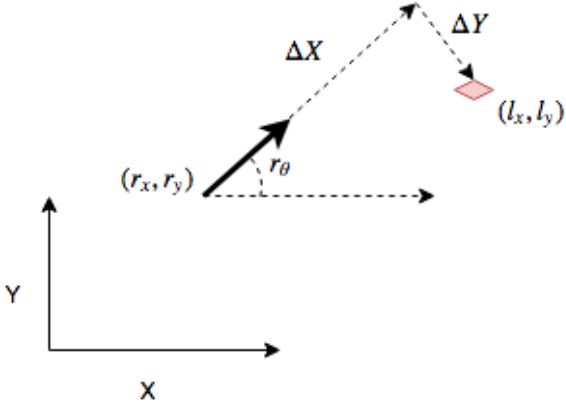}
    \caption{Frame visualization of sensor measurement}
    \label{fig: landmark}
\end{figure}
For sensor measurements, GroundsBot reads the relative pose of the AprilTag landmarks.  Because we only care about relative x and y position of the landmark, our sensor measurement takes the following form (equation \ref{eq:meassetup}) where $\Delta X$ is the relative x position of the landmark and $\Delta Y$ is the relative y position of the landmark. The global position of a landmark is described in equation \ref{eq:sensortolandmark}. From that, the predicted state change can be calculated as shown in equation \ref{eq:measurementfunction}. Finally, the measurement Jacobian is symbolically presented in equation \ref{eq:sensorjac}, and in full in Appendix  \ref{appendix:sensormodel}.

\begin{equation}
\label{eq:meassetup}
z =
\begin{bmatrix}
\Delta x\\
\Delta y\\
\end{bmatrix}
\end{equation}

\begin{equation}
\label{eq:sensortolandmark}
\begin{bmatrix}
l_x\\ 
l_y
\end{bmatrix} = \begin{bmatrix}
r_x + \Delta x cos(r_\theta) + \Delta y sin(r_\theta)\\ 
r_x + \Delta x sin(r_\theta) - \Delta y cos(r_\theta)
\end{bmatrix}
\end{equation}

\begin{equation}
\label{eq:measurementfunction}
\begin{bmatrix}
\Delta x\\
\Delta y 
\end{bmatrix} = 
\begin{bmatrix}
cos(r_\theta)(l_x - r_x) + sin(r_\theta)(l_y - r_y) \\
sin(r_\theta)(l_x - r_x) - cos(r_\theta)(l_y - r_y)
\end{bmatrix}
\end{equation}

\begin{equation}
\label{eq:sensorjac}
H=
\begin{bmatrix}
\frac{ \partial \Delta x} { \partial r_{x}} & \frac{ \partial \Delta x} { \partial r_{y}} & 
\frac{ \partial \Delta x} { \partial r_{\theta}} & \frac{ \partial \Delta x} { \partial l_{x}} & 
\frac{ \partial \Delta x} { \partial l_{y}}
\\
\frac{ \partial \Delta y} { \partial r_{x}} & \frac{ \partial \Delta y} { \partial r_{y}} & 
\frac{ \partial \Delta y} { \partial r_{\theta}} & \frac{ \partial \Delta y} { \partial l_{x}} & 
\frac{ \partial \Delta y} { \partial l_{y}}
\end{bmatrix}
\end{equation}

\subsection{Solver Infrastructure and Code}
To create a map, gradient descent was performed on the Gauss-Newton formulation to iteratively reduce the error of a state vector containing landmark positions and robot poses. At each iteration, an optimal state change vector $\Delta^{*}$ will be formulated. $\Delta^{*}$ is described in equation \ref{eq:lsq}.

\begin{equation} \label{eq:lsq}
\Delta ^  {*} =\underset{\Delta}{\mathrm{argmin}} 
\sum_{i}
\norm {
H_{i} \Delta_{i} - ( z_{i} - h_{i} (X_{i}^{0} ) )
}^{2}_{\Sigma_{i}}
\end{equation}
where $X_{i}^{0}$ is the state estimate at the current iteration,  $z_{i}$ is a vector containing odometry and landmark measurements,  $h_{i}$ contains measurement predictions based on the current iterations' state estimate, $H_{i}$ is the Jacobian of the measurement and control functions with respect to the state vector, as described in the sensor and motion model sections.
  
For a given iteration, equation \ref{eq:lsq} is in a form solvable by linear least squares approach, such as Cholesky decomposition (equation \ref{eq:axb}).
\begin{equation}
\label{eq:axb}
A x = b \xrightarrow{} H_{i} \Delta_{i} = z_{i} - h_{i} ( X_{i}^{0})
\end{equation}

\subsection {Estimating the Robots' Priors}

Based on the above process we have local maps generated for each robot.  We can now spatially merge the local maps by temporally concatenating the measurements from each robot to form a single global map using equation \ref{eq:generic_lsq}.  This final merging requires a the prior of each robot in a global frame.

One solution to find these priors is to define a single robot's frame as the global frame, determining the transform from this robot's local map to each other robot's local map using landmark correspondences, and then estimating each other robot's respective prior in the first robot's frame.

A strategy for finding the transform between two robots' local maps is shown in \cite{ddf-sam}.  Given a landmark correspondence $(l_i^{j}, l_i^{k})$ between landmark $l_i^j$ in the $j^{th}$ local map and its counterpart $l_i^k$ in the $k^{th}$ local, we define the relationship between an SE(2) transformation $\phi_k^j$ and the landmark correspondences as $l_i^{j}=\phi_k^j l_i^{k}$.

The correspondences needed to estimate the transform are given directly via AprilTag IDs. A finite number of points from the set of correspondences are chosen from each map. The relative rotation between the maps can be calculated with equation \ref{eq:relrot}.

\begin{equation}
\label{eq:relrot}
\theta_2^1 = 
tan^{-1} ( \frac {l_{b,y}^1 - l_{a,y}^1} {l_{b,x}^1 - l_{a,x}^1}) - 
tan^{-1} ( \frac {l_{b,y}^2 - l_{a,y}^2} {l_{b,x}^2 - l_{a,x}^2})
\end{equation}

After computing this relative rotation, the same can be used to estimate the translation between the map frames.

\begin{equation}
\label{eq:reltrans}
\mathbf{t_2^1 = r_1 - R_2^1 * r_2}
\end{equation}

where

$$
R_2^1=
\begin{bmatrix}
cos(\theta_2^1) & -sin(\theta_2^1) & 0 \\
sin(\theta_2^1) &  cos(\theta_2^1) & 0\\
0 & 0 & 1\\
\end{bmatrix}
,\ \ 
\mathbf{r_i} = 
\begin{bmatrix}
r_{ix} \\
r_{iy} \\
1
\end{bmatrix}
$$

The relevance of this estimation is dependent on the choice of correspondences. In order to account for inaccuracies in an estimated transformation and without the loss of generality, Random Sample Consensus \texttt{RANSAC} is used to find the set of landmark correspondences that minimizes their Euclidean distance after transformation.

\subsection { Solving for a Global Map }
With the prior of the second robot estimated in the frame first robot's, we can now solve for final global map.  To solve for a global, combined map, robot 2's map is appended to robot 1's map, with a prior set for robot 2's origin using the RANSAC transform estimate described above. The same non-linear solving technique used for each individual robot is then applied. Note that since measurements are relative to the robot-centric frames (not robot map frames), the only use of the transform is for robot 2's origin prior.

Robot 2's prior constraint is added as shown in equation \ref{eq:r2prior}, where $T = \begin{bmatrix} T_{x}& T_{y}& T_{\theta} \end{bmatrix}^{T}$ is the RANSAC prior, $x = \begin{bmatrix} r_{x}& r_{y}& r_{\theta}\end{bmatrix}$ is the current state estimate of robot 2's origin, and  $\frac { \partial h_{0}^{2} }  { \partial x }$ is the odometry Jacobian. Note the odometry Jacobian is the same as equation \ref{eq:odomjac}, with the exception that it uses robot 2's origin and robot 1's origin instead of sequential poses.

\begin{equation}
\label {eq:r2prior}
\frac { \partial h_{0}^{2} }  { \partial x } = 
\begin{bmatrix}
T_{x} \\ T_{y} \\ T_{\theta}
\end{bmatrix}
-
\begin{bmatrix}
r_x\\
r_y\\
r_\theta\\
\end{bmatrix}
\end{equation}

\section{Implementation} \label{implementation}
\subsection{Robot}
For the purposes of data acquisition and testing, we are working with the unmanned ground vehicle, GroundsBot.

\begin{figure}
    \centering
    \includegraphics[width=\linewidth]{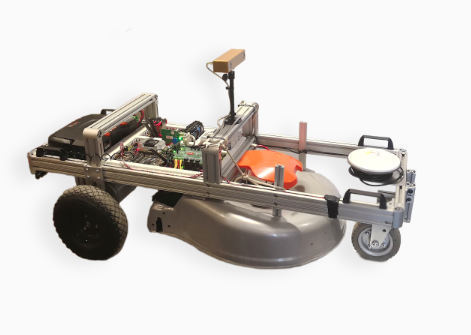}
    \caption{GroundsBot was developed to autonomously mow expansive lawns such as golf courses, soccer fields etc.}
    \label{fig: grudsby}
\end{figure}

GroundsBot is a mobile robot designed for mowing grass at golf courses, but can act as a robust and flexible platform for research.  It is useful for our project because it has an onboard camera for AprilTag recognition and wheel encoders for an odometry output.  Figure \ref{fig: grudsby} shows an image of the GroundsBot platform.

Specifically for landmark detection, we will be using the ROS package AprilTags open-sourced by Wang and Olson (2016) \cite{wang2016iros} to identify and localize our AprilTag landmarks.  The package takes in camera data and outputs the relative pose of the AprilTag from the camera's perspective.  For our odometry we will be using GroundsBot's wheel encoders to output the wheel velocities of the left and right differential wheels.

One challenge for our specific project however is the fact we only have one GroundsBot.  In order to do multi-robot mapping, we had GroundsBot traverse several different trajectories - which allowed us to incorporate the spatial and temporal changes in a multi-robot setting.  Each robot trajectory represents a set of states of the robot that will be merged into the final global map.
\subsection{Data}
Our end goal was to generate one single map of the first floor of Newell-Simon Hall by building two separate but overlapping maps of the floor.

We decided to use AprilTags for our landmark features to focus the scope of the project.  As specified in \cite{Dellaert06ijrr}, for good smoothing and mapping results, uniquely identifiable landmarks are required.  Instead of using computer vision techniques to detect and localize unique identifiers in a scene, we use AprilTags as our landmark features.  This allows us to easily determine the id of a landmark and the distance to a landmark.  For the control input we will be using wheel velocities as our odometry measurements. The measurement and control inputs are then used to build the least squares representation of the robots' trajectories and sensor measurements.

The final test scenario had GroundsBot navigate two separate paths and use the AprilTags as features. The intended scenario is shown below in figure \ref{fig: scenario}.

\begin{figure}[b]
    \centering
    \includegraphics[width=\linewidth]{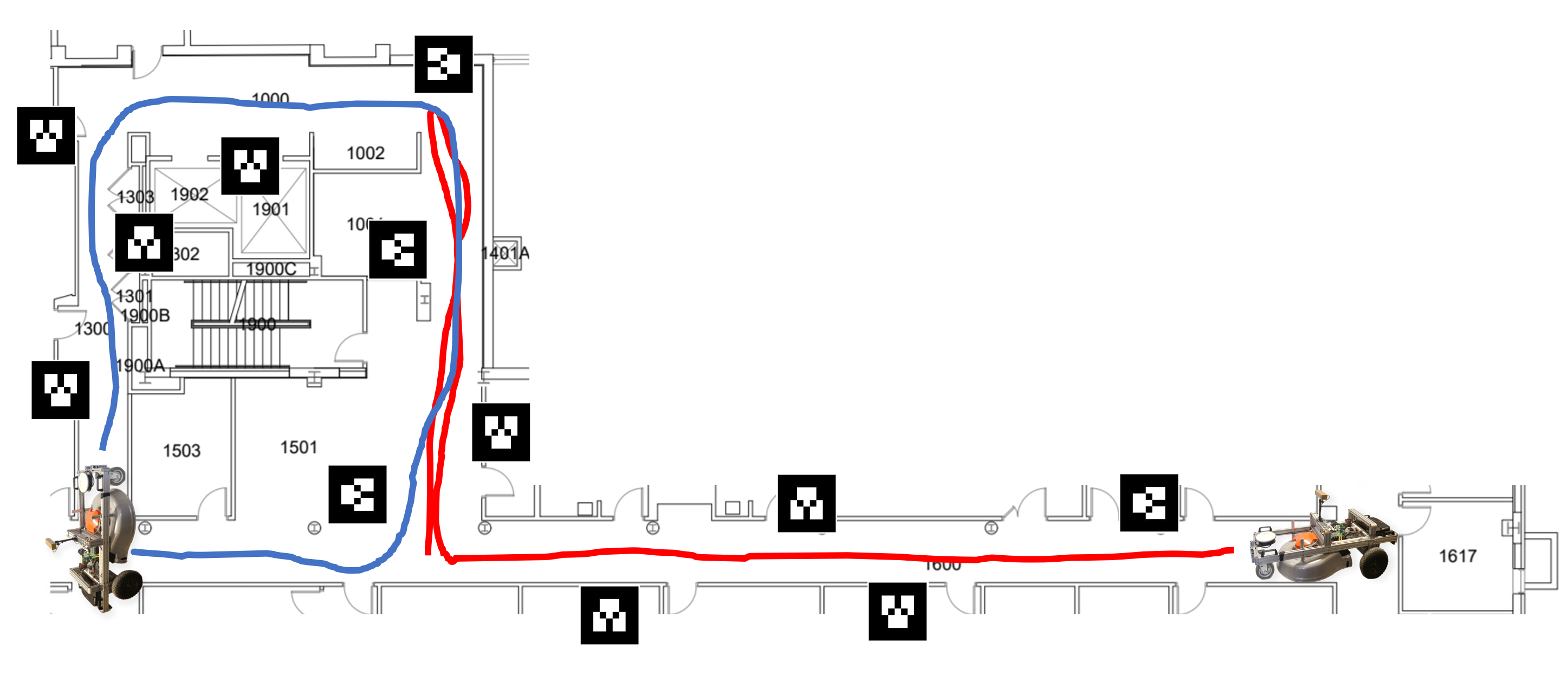}
    \caption{Goal scenario with two separate but overlapping paths}
    \label{fig: scenario}
\end{figure}

\begin{figure}
    \centering
    \includegraphics[scale=.04]{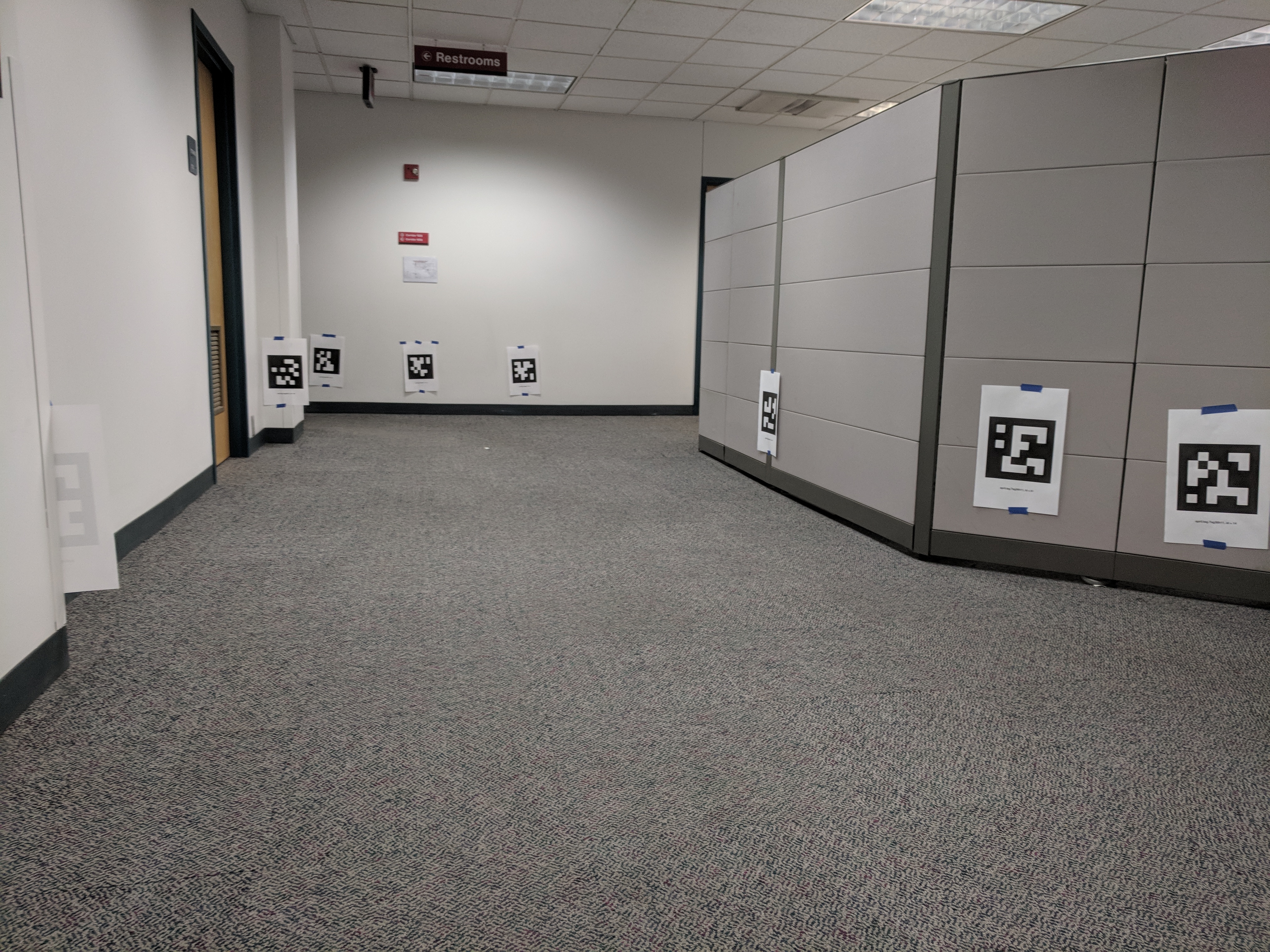}
    \caption{Test setup on first floor}
    \label{fig: testsetup}
\end{figure}

For the test set up, we distributed fifty AprilTags around the first floor of NSH.  Figure \ref{fig: testsetup} shows an example on how the AprilTags were distributed.  GroundsBot was then manually navigated around the center loop of the first floor to create a trajectory to be used for the first local map. Next, GroundsBot was teleoperated down the hallway and around part of the center loop to create the trajectory for the second local map. ROS bag files were collected from both these runs and were used to provide the necessary odometry and landmark measurement for the map generation.

To process the bag files, we created a post-processing script to extract the odometry and landmark measurements of the trajectory as well as synchronize the odometry readings with the landmark measurements. This script outputs data to a csv file that can be parsed by our non-linear SLAM solver.

\section{Results}
\subsection{Local Map Generation}
\begin{figure}[t]
\begin{minipage}{0.47\textwidth}
    \begin{center}
    \includegraphics[scale=.55]{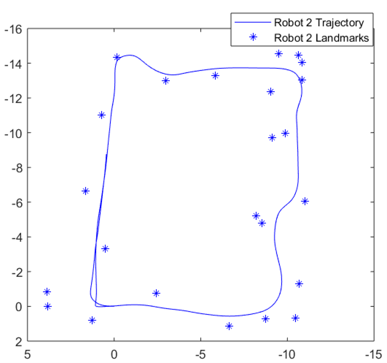}
    \end{center}
    \caption{Map of Loop Section of NSH Floor 1}
    \label{fig: loop}
    \end{minipage}
    \hspace{\fill} 
    \begin{minipage}{0.47\textwidth}
    \includegraphics[width=\linewidth]{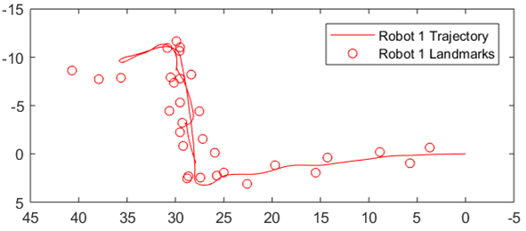}
    \caption{Map of Hallway Section of NSH Floor 1}
    \label{fig: hallway}
    \end{minipage}
\end{figure}

\begin{figure}[t]
    \includegraphics[width=\linewidth]{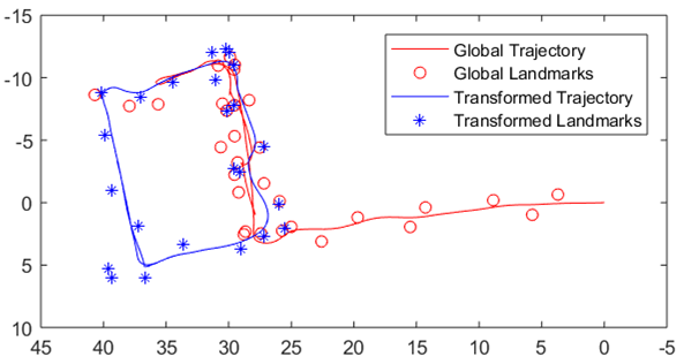}
    \caption{Individual Maps Superimposed After Estimating Transformation}
    \label{fig: ransac}
\end{figure}

\begin{figure}[t]
    \includegraphics[width=\linewidth]{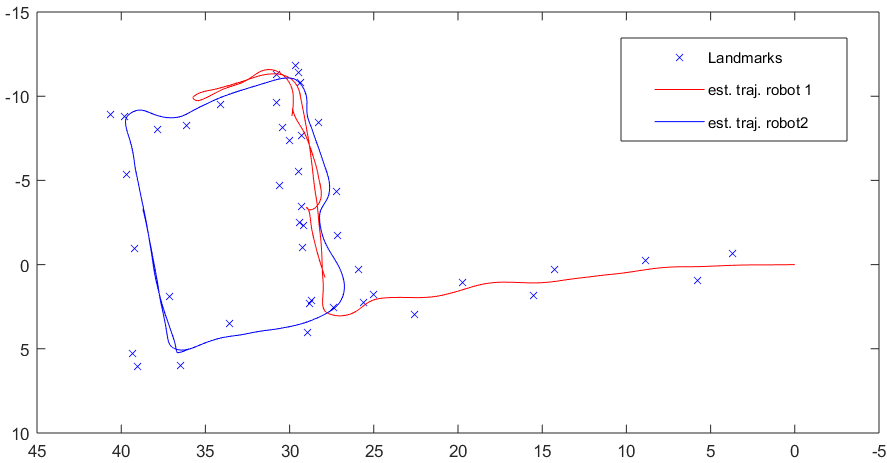}
    \caption{Full Bundle Adjustment Post RANSAC}
    \label{fig: combined}
\end{figure}

The results of the independent local map from each of the two runs can be seen in figures \ref{fig: loop} and \ref{fig: hallway}. Figure \ref{fig: loop} shows the resulting map after looping GroundsBot around the floor. This map shows accurate loop closure once GroundsBot had driven fully around the loop. Figure \ref{fig: hallway} shows the resulting map after driving GroundsBot down the hallway and overlapping with the loop route. In both images a small amount of drift can be seen but the overall trajectory of each robot is easily recognizable as a part of NSH floor 1.
\subsection{RANSAC Transformation}
Figure \ref{fig: ransac} shows the results of naive transformation of the map of the \texttt{loop} (\ref{fig: loop}) into the \texttt{hallway} maps (\ref{fig: hallway}) frame. Ideally, all landmarks being represented as a '*' should land directly in the middle of their corresponding 'O' landmarks. This does happen for a few of the landmarks but there is some amount of error in the rest of the landmarks. It is worth noting that there are a few instances of landmarks that look to be corresponding but in reality are distinct landmarks (e.g. two '*' in the top right corner of the loop). This is further clarified in the final combined map described in the next section. 
\subsection{Global Map Generation}
As part of the smoothing and global map generation, a final bundle-adjustment is performed. The generated global map shown in figure \ref{fig: combined}.  Notice each landmark is only represented once in the final global map.  We will also observe some landmarks move more than others.  This is a function of not only how close the two robots viewed the two landmarks initially, but also how many measurements of a landmark one robot received compared to the other robot.  If one robot detects a landmark in relatively higher number of independent measurements than the other robot, the final landmark position will be a weighted average favoring the former robot's estimated position.
To quantitatively assess the success of the map merging, a ground truth distance from robot 1's starting position to robot 2's starting position was manually measured during data collection to be 38.05 meters. That distance on the merged map was found to be 37.37 meters, yielding an error of 1.8 percent.






\bibliographystyle{plainnat}
\bibliography{references}

\newpage
\begin{appendices}
\section{Sensor Model Jacobian}
\label{appendix:sensormodel}

$$
H=
\begin{bmatrix}
\frac{ \partial \Delta x} { \partial r_{x}} & \frac{ \partial \Delta x} { \partial r_{y}} & 
\frac{ \partial \Delta x} { \partial r_{\theta}} & \frac{ \partial \Delta x} { \partial l_{x}} & 
\frac{ \partial \Delta x} { \partial l_{y}}
\\
\frac{ \partial \Delta y} { \partial r_{x}} & \frac{ \partial \Delta y} { \partial r_{y}} & 
\frac{ \partial \Delta y} { \partial r_{\theta}} & \frac{ \partial \Delta y} { \partial l_{x}} & 
\frac{ \partial \Delta y} { \partial l_{y}}
\end{bmatrix}
$$

$$
\frac{ \partial \Delta x} { \partial r_{x}} = -cos(r_{\theta}) 
$$
$$
\frac{ \partial \Delta x} { \partial r_{y}}  = -sin(r_{\theta})  
$$
$$
\frac{ \partial \Delta x} { \partial r_{\theta}} = -sin(r_{\theta})(l_x - r_x) + cos(r_{\theta})(l_y - r_y)  
$$
$$
\frac{ \partial \Delta x} { \partial l_{x}} = cos(r_{\theta})  
$$
$$
\frac{ \partial \Delta x} { \partial l_{y}} = sin(r_{\theta})
$$

$$
\frac{ \partial \Delta y} { \partial r_{x}}= -sin(r_{\theta}) 
$$
$$
\frac{ \partial \Delta y} { \partial r_{y}} = cos(r_{\theta})  
$$
$$
\frac{ \partial \Delta y} { \partial r_{\theta}} = cos(r_{\theta})(l_x - r_x) + sin(\theta)(l_y - r_y)  
$$
$$
\frac{ \partial \Delta y} { \partial l_{x}} = sin(r_{\theta})  
$$
$$
\frac{ \partial \Delta y} { \partial l_{y}} = -cos(r_{\theta})
$$

\section{Challenges}
The team faced several different challenges throughout the implementation of this project. One such challenge was around data collection. It took multiple trials to get the position of the AprilTags correct so as to be consistently registered. Initially, the AprilTags were spread out somewhat randomly in the basement of NSH. After our initial collection and analysis, we realized we needed to be much more deliberate about where we placed the AprilTags so that the area we were mapping was recognizable against available ground truth/ floor plans and verifiable based on it's general layout.

Another challenge was dealing with the data output of GroundsBot. There were two main issues with regards to the output. First, the odometry and landmark data was not synced. To resolve this a state counter was added and incremented whenever a new odometry message was read and published. When a new landmark measurement is received, the appropriate state is added to the measurement by checking the current value of the state counter. The second issue was the rate of the two data streams. Odometry data is published substantially more often than the landmark measurements resulting in a large number of robot states that have no associated measurements. To account for this, we subsampled every fifth odometry measurement.

We also faced challenges in integrating the second robot's origin prior. Conceptually, the methods used in  \cite{ddf-sam} are straightforward. In practice, modifying our non-linear solver to produce the constraint on robot 2's origin proved difficult to test and debug. We developed a simulation script to generate the odometry and location measurements for two robots, and were then able to step through the solver one iteration at a time to diagnose and resolve indexing errors.

\end{appendices}
\end{document}